# Segmentation of Bleeding Regions in Wireless Capsule Endoscopy for Detection of Informative Frames

M. Hajabdollahi[1], R. Esfandiarpoor[1], P. Khadivi[2], S.M.R. Soroushmehr[3,4], N. Karimi[1],
K. Najarian[3,4,5], S. Samavi[1,3]

[1]Department of Electrical and Computer Engineering, Isfahan University of Technology, Isfahan 84156-83111, Iran.
[2]Department of Computer Science, Seattle University, Seattle, 98122 USA.
[3]Michigan Center for Integrative Research in Critical Care, University of Michigan, Ann Arbor, 48109 U.S.A.
[4]Department of Computational Medicine and Bioinformatics, University of Michigan, Ann Arbor, 48109 U.S.A.
[5]Emergency Medicine Department, University of Michigan, Ann Arbor, 48109 U.S.A.

*Abstract*— **Wireless capsule endoscopy (WCE) is an effective mean for diagnosis of gastrointestinal disorders. Detection of informative scenes in WCE video could reduce the length of transmitted videos and help the diagnosis procedure. In this paper, we investigate the problem of simplification of neural networks for automatic bleeding region detection inside capsule endoscopy device. Suitable color channels are selected as neural networks inputs, and image classification is conducted using a multi-layer perceptron (MLP) and a convolutional neural network (CNN) separately. Both CNN and MLP structures are simplified to reduce the number of computational operations. Performances of two simplified networks are evaluated on a WCE bleeding image dataset using the DICE score. Simulation results show that applying simplification methods on both MLP and CNN structures reduces the number of computational operations significantly with AUC greater than 0.97. Although CNN performs better in comparison with simplified MLP, the simplified MLP segments bleeding regions with a significantly smaller number of computational operations. Concerning the importance of having a simple structure or a more accurate model, each of the designed structures could be selected for inside capsule implementation.**

## I. INTRODUCTION

Wireless capsule endoscopy (WCE) is a non-invasive, painless endoscopic method employed in screening different parts of the gastrointestinal (GI) organs which is used for a variety of medical experiments [1]. WCE imaging is always used for patients suspicious of bleeding and other types of abnormalities such as Crohn's disease in GI. Also, it is possible to use WCE for patients with polyposis syndrome and small bowel disorder [1]. Researchers have investigated automatic methods for detection of abnormalities since 2001, the time that WCE was first launched [1]. These methods have drastically reduced the time spent by a physician for medical diagnostic and detection [2].

In recent years, researchers have developed different methods for automatic detection of a variety of abnormalities such as ulcer, bleeding, polyps and other abnormalities in WCE images. In some studies such as [3] and [4], handcrafted features are employed. In [3], saliency is calculated based on color and texture information. For texture representation LOG-Gabor filter, *scale invariant feature transform* (SIFT) and *local binary pattern* (LBP) are used. For color information, second channels of HSI and CMYK color spaces are used. All features are coded with K-means clustering, and final saliency map is obtained by max-pooling of the saliency map and coded features. Mehmood *et al.* used different-order of moments as a saliency analysis method to summarize WCE video frames [4].

Machine learning methods have gained a lot of attention in recent years due to their remarkable success in medical image processing applications. Among them, support vector machine (SVM) is used extensively for abnormality detection in different human organs. Fu *et al.* [5], used super pixel color features to speed up the bleeding detection process and employed SVM as a classifier. Suman *et al.* [6], removed noise, edges, dark, and light regions to avoid misclassification of bleeding regions. In their work, a combination of RGB channels at the block level is used as SVM input. In [7], a K-means algorithm is employed to cluster bleeding images into two clusters based on the block's statistical characteristics. Bleeding images are then detected from non-bleeding images using differential features, and finally, bleeding zones are localized using an SVM.

In many studies, different features such as SIFT, LBP, *speed up robust feature* (SURF), Gabor filter, the *histogram of the gradient* (HOG) and *discrete wavelet transform* (DWT) are used as SVM classifier's inputs [8-14]. In [8], LBP feature is extracted from the "*I*" channel of the HSI color space in 8×8 image blocks and then each block is segmented by SVM. Deeba *et al.* [9], presented a method based on cellular automata for clustering WCE images. For segmentation of bleeding regions, they utilized SVM to classify the cluster centroids. In [10], the histogram of pixels is assigned to each cluster in YCbCr color space, then it is used as a descriptor, and by using an SVM, WCE images are segmented. In [11], a method for detection of polyp and ulcer in WCE frames is presented where Gabor filter is applied for simulating the human visual system. Also, edges are detected using *smallest univalue segment assimilating nucleus* (SUSAN) edge detection method to improve the accuracy of the detection algorithm. Then, using the results from previous stages, regions containing ulcer and polyp are detected base on SVM and a decision tree. Yuan *et al.* [12], extracted different features around image key points including LBP, SIFT, and

HOG. In their work, the set of extracted features are clustered and classified using an SVM classifier. In [13], a method for detection of abnormalities in WCE images is presented where color and texture features are extracted using moments, LBP and DWT. Importance of the features is identified using a saliency map and based on this importance, WCE images are classified by SVM. In [14], a method for automatic lesion detection in capsule endoscopy is presented. After CIE-Lab color transformation, salient points are detected via SURF detector based on the area around each pixel. Using features of salient points, SVM is employed for classification of regions to lesion and non-lesion.

Recently, deep neural networks achieved considerable successes in the detection and segmentation of abnormalities in WCE images. Jia *et al.* [15], proposed a bleeding frame detection method consisting of feature extraction and classification steps. In their work, histograms of K-Means clustering are used as input features and are classified by a *convolutional neural network* (CNN). Also, Jia *et al.* [16], augmented datasets by rotation and mirroring to extend the number of training and test objects and used a CNN structure for detection of bleeding frames. In [17], bleeding images are classified into bleeding and non-bleeding using SVM and then in each classified image, bleeding regions are segmented using CNN. In [18], image patches are fed into a single CNN structure and classified as normal and abnormal.

In WCE videos, only a few frames are informative and transmitting all the frames will increase the battery usage of WCE devices. Hence, designing a low power system for image analysis and informative frame detection is very demanding due to limited memory capacity and limited power supply [4]. Also, there is a great demand for the implementation of the diagnostic process inside the WCE device. Although in some previous works powerful tools such as CNN were utilized for segmentation, the complexity of the detection method can be problematic for inside capsule implementation. For example, the method of [16], uses CNN with 15M parameters and method in [17], uses a fully-convolutional network with 6M parameters [19]. A large number of parameters and the use of *multiply-accumulate* (MAC) operations make the use of CNNs infeasible considering the size and resources of the capsule endoscopy.

Several methods have been developed for hardware implementation of image processing algorithms inside the capsule device. In [20], a hardware architecture for WCE image assessment inside the capsule is presented. Informative frames are detected inside the capsule using mean, variance, skewness and kurtosis. Image compression inside WCE is introduced as another way to reduce the power consumption needed for transferring the video frames to outside. In [21], a hardware architecture for image compression inside WCE is proposed. To compress the images, an integer based DCT, and an efficient coefficient encoding with low complexity is applied. In [22], a hardware core including a camera interface, *first in first out* (FIFO) queue, controller, and image compressor is presented. In [23], an architecture for calculation of DCT transform which is employed in image compression inside WCE is designed. The applied DCT is a 1-D DCT transform which reduces the number of addition and shift operations.

In this paper, we utilize neural networks including a CNN and an MLP as complex and simple structures respectively for bleeding region detection in WCE images. Different color spaces are analyzed to select channels containing more information about bleeding regions. To have a simple and efficient network structure, different network simplification approaches are investigated. In case of the MLP structure by quantization approach multiplications are removed, and in case of the CNN by a simultaneous quantization and pruning approach, multiplications are removed in *fully connected layers* (FCLs) and redundant computations in *convolutional layers* (CLs) are eliminated. Both network structures can segment bleeding regions with high DICE score. In both proposed network structures for bleeding region segmentation, computational complexity is reduced significantly, and this reduction provides an opportunity to implement the proposed methods inside the capsule device. The main contributions of this study are summarized as followings.

- Selection of proper color channels based on mutual information as inputs of utilized neural networks.
- Designing a simplified MLP structure suitable for image analysis inside the capsule endoscopy device.
- Employing simultaneous quantization and pruning to simplify the CNN structure with a little drop of accuracy.

The remainder of this paper is organized as follows. In Section II, neural network simplification techniques are briefly explained. In Section III, the proposed method for the segmentation of bleeding regions in WCE images is demonstrated. Section IV is dedicated to experimental results and finally concluding remarks are presented in Section V.

## II. Neural network simplification techniques

Machine learning techniques are employed for medical image analysis in various research projects conducted in recent years. Among different machine learning methods, *artificial neural networks* (ANNs) have considerable achievements in automatic segmentation and classification of a variety of abnormalities [24-26]. CNNs, as an advanced version of primary neural networks, can appropriately accomplish the classification and segmentation tasks. Although CNNs are usually more accurate than simple MLP networks and have better capabilities in automatic feature selection, their structures are more complex than MLP. When it comes to implementing CNN on devices with limited computational and power resources, their structural complexity is forbidding. Different techniques were proposed to alleviate the complexity of neural network structures. These methods are briefly explained in the following subsections.

### A. Neural Network quantization

In an ANN, computational operations are mainly due to three categories including multiplication of weight matrices, addition of activation function inputs and calculation of activation function outputs. The number of computational operations can be reduced in different ways. Recently, binarization is used as an efficient way to reduce the complexity of the networks' structure [27-28]. During binarization, all of the weights are affected by the simplification process. In the binarization process, a value is converted to two possible values, as an example to 0 and 1. Two methods were introduced in previous studies for binarization [27-28]. The first approach is a deterministic binarization illustrated in the following equation:

$$W_b = \begin{cases} +1 & if\ W \geq 0, \\ -1 & otherwise. \end{cases} \quad (1)$$

where $W$ is a network weight before quantization and $W_b$ is the binarized one. The second approach for binarization is the stochastic method shown in the following equation:

$$W_b = \begin{cases} +1 & with\ probability\ p = \sigma(w), \\ -1 & with\ probability\ 1-p. \end{cases} \quad (2)$$

where $\sigma(x)$ is "hard sigmoid function" defined as follows:

$$\sigma(x) = clip\left(\frac{x+1}{2}, 0, 1\right) = max(0, min(1, \frac{x+1}{2})) \quad (3)$$

The binarization could decrease WCE power consumption. In [29], the power consumption of the operations with different representations was roughly presented. As shown in Table I, the energy consumption of addition and multiplication operations is reduced significantly with a shorter bit width of the computational operations [29]. Also, concerning [29], binary representation consumes 32 times smaller memory size and 32 times fewer memory accesses in comparison with 32-bit representation.

Table I. Rough energy costs for various operations in 45 nm [29]

| Operation | Multiply | Add |
|---|---|---|
| 8-bit Integer | 0.2pJ | 0.03pJ |
| 32-bit Integer | 3.1pJ | 0.1pJ |
| 16-bit Floating Point | 1.1pJ | 0.4pJ |
| 32-bit Floating Point | 3.7pJ | 0.9pJ |

*B. Neural network pruning*

After training neural networks, there might be some redundant weights that can be removed or merged with other weights without significant degradation in network accuracy. By removing redundant weights, the problem of redundant computations is alleviated especially for complex neural networks. Also, network pruning can be used to address the over-fitting problem while training neural networks. Different methods have been proposed for pruning such as weight decay [30], and Hessian of the loss function [31]. However, their proposed pruning techniques are not very efficient and in terms of computational complexity second order derivatives are required. In [32], a magnitude-based pruning was developed which affects weights and connections individually. In [33], a filter pruning method is presented in which all coefficients of a filter are removed. Although in filter pruning better structural sparsity is obtained, accuracy drop due to the pruning of a filter can be substantial. Therefore, it cannot be used as a pruning method in general. Principally, all of the pruning methods consist of three steps. In the first step network variables are trained to achieve acceptable accuracy. In the second step, redundant weights are removed. Finally, to compensate for the reduction of the accuracy, caused by the operations of the second step, the network is retrained. In the primary pruning method, weights are removed during the training process based on a pruning rate, and finally, a pruning mask represents which weights are removed. Pruning mask is calculated using Eq. (4) in which $W_l$ and $Mask_l$ are network weight matrix and pruning mask matrix in layer $l$ respectively and $\alpha$ is the pruning rate.

$$Mask_l(i,j) = \begin{cases} 0 & W_l(i,j) < \alpha * \sigma^2(W_l) \\ 1 & Otherwise \end{cases} \quad (4)$$

As shown in (4), a sample weight in layer $l$ as $W_l(i,j)$ is pruned if its value is less than a threshold. The threshold is equal to the pruning rate multiplied by the standard deviation of all weights in layer $l$ (i.e. $\sigma^2(W_l)$). Pruning rate can be modified during the training process to get better results. Finally, pruning is performed during back propagation as follows:

$$W(i,j) = W(i,j) - \eta \left(\frac{\partial C}{\partial W(i,j)} * Mask_l(i,j)\right) \quad (5)$$

In (5), weights are updated by a learning rate $\eta$ and gradients of the weights with small values are removed from the training process by multiplying the gradients matrix by $Mask_l$. Network error in terms of $W(i,j)$ is specified by $\frac{\partial C}{\partial W(i,j)}$, which can be compensated by retraining [32].

III. SIMPLIFIED SEGMENTATION METHOD

The proposed method includes two major parts in general. The first part is dedicated to simplification of the MLP, and the other one explains the simplification of the CNN. Before applying the method to a neural network, a color space conversion is required to reach better performance. The proposed method is described in the following subsections in more details.

*A. Color space conversion*

Input images can be represented in different color spaces and in each color space, some features are more representative. As an example, in retinal image analysis, the green channel is more informative than the others [25], [34]. Color space mapping can be considered as a preprocessing step applied in medical image processing techniques. HSV and CIE-*lab* color spaces are two alternatives to RGB color space. CIE-*lab* is more intuitive than RGB and is designed to approximate the human vision system [35]. "$l$," "$a$" and "$b$" are components of the CIE-*lab* so that "$l$" lies in 0 to 100 and both "$a$" and "$b$" lie in $-110$ to $110$. HSV color space including hue, saturation, and value is quite similar to the human's perception of colors. In Fig. 1, a bleeding frame of WCE videos is shown in different color spaces. As illustrated in Fig. 1, some color spaces represent bleeding regions better than the others, and some of them have no useful meaning. From a visual perspective, Fig. 1 shows that the gray-scale domain, the saturation channel, and the "$a$" channel have distinctive representations of the bleeding region. Using an experiment based on information theory, it is possible to identify color spaces or channels which are suitable for representation of the bleeding regions.

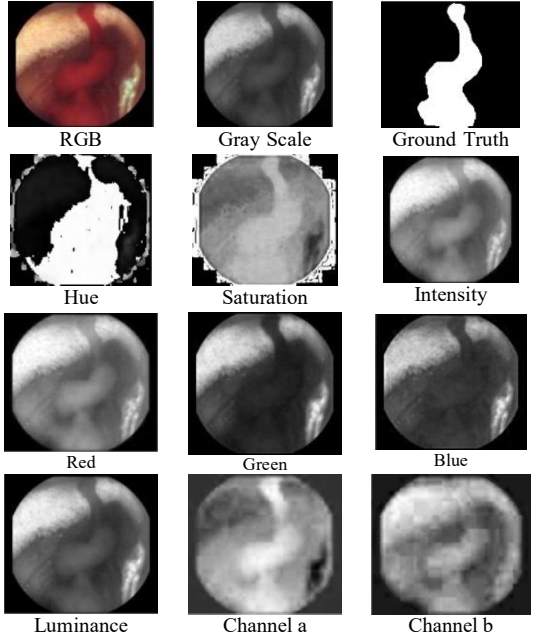

Figure 1. Different color space in a sample WCE image

*1) Color space and channel selection based on mutual information*

Mutual information is the amount of information that one random variable contains about another random variable and can be calculated using the following equation [36].

$$I(X;Y) = H(X) + H(Y) - H(Y;X) \qquad (6)$$

In (6), $X$ and $Y$ are random variables and $H$ is the entropy function. The mutual information between each color channel and the ground truth can be calculated for different color spaces, as a measure of the applicability of the channels. For 55 WCE images in [37] and [38], 10 color spaces and channels including "H", "S", and "V" in HSV color space, "$l$", "$a$", and "$b$" in CIE-*lab* color space, "R", "G", and "B" in RGB color space and grayscale level image are experimented. Experiments are conducted to detect single channels, 2-combinations, and 3-combinations of color channels with the most information about ground truth. In the first experiment, the mutual information of each 10 transformed images and their corresponding ground truth is calculated using Eq. (6), and the results are illustrated in Fig. 2. It can be observed that the three most informative color channels are "$a$", "G", and "S". Also, to identify the most informative pair of color channels, all possible pairs of 10 color channels are considered. The latter experiment resulted in 45 states illustrated as a matrix in Fig. 3 where the least three informative 2-combinations of color channels are shown in red. The three most informative pairs of color channels can be considered as ($a$, B), ($a$, S), and ($a$, Gray) which are shown in green in Fig. 3. To identify the most informative 3-combinations of color channels, all possible 3-combinations of 10 color channels are considered. This experiment resulted in 120 states. According to this experiment, the three least informative 3-combinations of color channels are (R, H, V), (R, V, Gray), and ($l$, R, V). Moreover, the three most informative 3-combinations of color channels are ($a$, $l$, S), ($a$, G, S), and ($a$, S, Gray).

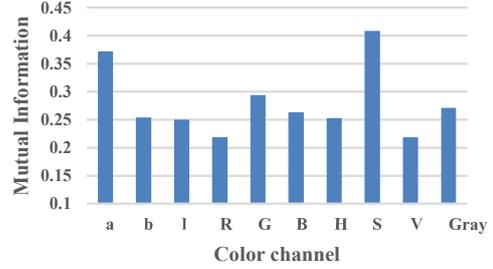

Figure 2. Mutual information corresponding to single color channels

|   | *a* | *b* | *l* | R | G | B | H | S | V | Gray |
|---|---|---|---|---|---|---|---|---|---|---|
| *a* | 0.37 | 0.40 | 0.48 | 0.48 | 0.49 | 0.50 | 0.40 | 0.49 | 0.48 | 0.49 |
| *b* | 0.40 | 0.25 | 0.40 | 0.37 | 0.46 | 0.44 | 0.39 | 0.45 | 0.37 | 0.42 |
| *l* | 0.48 | 0.40 | 0.25 | 0.33 | 0.45 | 0.42 | 0.31 | 0.45 | 0.33 | 0.45 |
| R | 0.48 | 0.37 | 0.33 | 0.22 | 0.40 | 0.43 | 0.30 | 0.45 | 0.22 | 0.36 |
| G | 0.49 | 0.46 | 0.45 | 0.40 | 0.29 | 0.39 | 0.36 | 0.45 | 0.40 | 0.43 |
| B | 0.50 | 0.44 | 0.42 | 0.43 | 0.39 | 0.26 | 0.39 | 0.42 | 0.43 | 0.41 |
| H | 0.40 | 0.39 | 0.31 | 0.30 | 0.36 | 0.39 | 0.25 | 0.47 | 0.30 | 0.33 |
| S | 0.49 | 0.45 | 0.45 | 0.45 | 0.45 | 0.42 | 0.47 | 0.40 | 0.45 | 0.45 |
| V | 0.48 | 0.37 | 0.33 | 0.22 | 0.40 | 0.43 | 0.30 | 0.45 | 0.22 | 0.36 |
| Gray | 0.49 | 0.42 | 0.45 | 0.36 | 0.43 | 0.41 | 0.33 | 0.45 | 0.36 | 0.27 |

Figure. 3. Mutual information corresponding to 2-combinations of color channels

### B. Simplified MLP neural network

WCE devices suffer from limited computational capabilities and power resources and hence, simple MLP structures can be used more efficiently. As illustrated in Fig. 4, an MLP structure can be used for segmentation of the bleeding regions in WCE images. Three color channels containing the most information are extracted from a frame of WCE video. A patch is considered around each pixel, and extracted patches from the three color channels are aligned as inputs of the MLP. In the training phase, the class of the central pixel is considered as the output of the network for each patch. Two neurons are considered as the output of the network, and a softmax activation function indicates the final class of the input patch. Considering resource limitations of the WCE

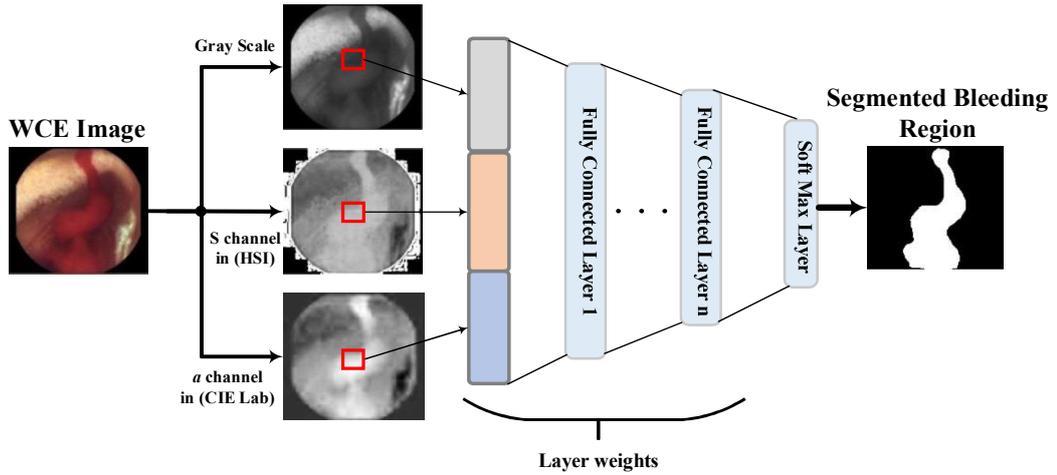

Figure 4. Overview of the MLP architecture

device, no further feature extraction and preprocessing have been applied.

Quantization and pruning are effective methods to reduce the number of computational operations required by neural networks. Quantization reduces the width of computational operations, and pruning reduces the number of computational operations.

For further simplification and preparation of the overall system for being embedded inside the capsule device, the network weights are quantized. In this paper, deterministic binary quantization of the weights is considered as follows:

$$W_q = \begin{cases} -1 & if\ W < 0 \\ 1 & if\ W \geq 0 \end{cases} \quad (7)$$

in which $W$ and $W_q$ are original and quantized network weights respectively.

Using this quantization, all multiplications are removed, and the heavy parts of the computations required for conducting segmentation are eliminated. Quantization algorithm is presented as the pseudo code in Fig. 5. In Fig. 5, N is the number of epochs, L is the number of network layers, $W_l$ is the weights matrix in layer $l$, $W_l^q$ is its corresponding quantized weight matrix which is in form of binary values, $C$

---

*Train network parameters for a few epochs*
**For** Epoch 1 to N
    **For** $l$=1 to L
        $W_l^q = Quantize\ (W_l)$
    **END**
    *Forward and backward path using* $W_l^q$
    **For** $l$=1 to L
        $W_l = W_l - \eta\ (\frac{\partial C}{W_l^q})$
    **END**
**END**

Figure. 5. Quantization algorithm for MLP

---

is the cost function, and $\eta$ is the learning rate. As illustrated in the first line of Fig. 5, to achieve better performance, in the first epochs of training, variables are not quantized. After that, variables are quantized during training and loss is calculated using $W_l^q$. After each quantization step, the network is retrained to recover its accuracy drop and finally, weights with real values are updated using gradients resulted from the back propagation step.

### C. CNN Quantization and Pruning

In most of the cases, problems arise during both pruning and quantization. While quantizing, bit length of variables reduces extensively which may result in problems in sections of the network with a small number of parameters. For example in convolutional layers, small numbers of coefficients are employed in each filter, and a major variation in filter coefficients due to the quantization leads to an undesirable effect on the network performance. Also, pruning has some problems due to the utilization of full precision operations. Although in pruning some connections are removed, the remaining multiplications consume a lot of computational units from a hardware perspective due to the use of weights with full bit length. Our proposed method addressed the problems above by employing the quantization and pruning of the CNN simultaneously. In Fig. 6, a CNN architecture for segmentation of bleeding regions is illustrated. This structure is employed for simultaneous pruning and quantization. Image patches are selected from three color channels and are used as inputs of the *convolutional layers* (CLs) which are followed by the *fully connected layers* (FCLs). The weights of the FCLs are quantized and to avoid the severe drop of accuracy due to the simplification of CLs, their weights are pruned. After each simplification phase to compensate for the drop in accuracy resulted from the simplification, network parameters are retrained. Details of the training algorithm using simultaneous quantization and pruning techniques are illustrated in Fig. 7. In this pseudo code N is the number of epochs, L is the number of network layers and $W_l$ is network

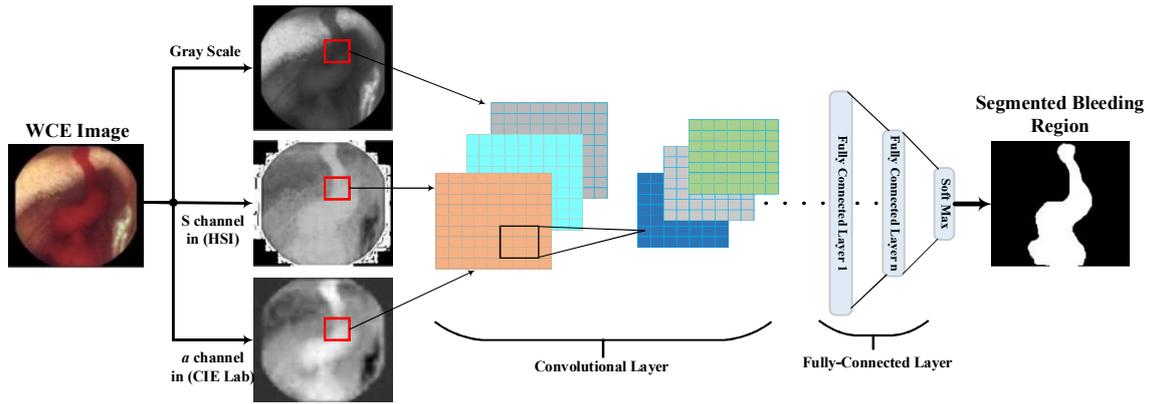

Fig. 6. Overview of the CNN architecture

```
Train network parameters for a few epochs
Calculate pruning mask (Mask_l)
 For Epoch 1 to N
   For l=1 to L
     IF W_l is a conv filter
        W_l = W_l * Mask_l
     ELSE W_l is a FCL layer
        W_l^q = quantize(W_l)
   END
   Forward and backward path using W_l and W_l^q for
   convolutional filters and FCLs, respectively
   For l=1 to L
     IF W_l is a conv filter
        W_l = W_l − η (∂C/∂W_l) ⊙ Mask_l
     Else W_l is a FCL layer
        W_l = W_l − η (∂C/∂W_l^q)
   END
 END
```

Figure. 7. Simplification algorithm for CNN

weights matrix in layer $l$, $W_l^q$ is its corresponding quantized weight matrix, $Mask_l$ is pruning mask in layer $l$, $\eta$ is learning rate, $C$ is cost function and $\odot$ is the Hadamard production operation. At first, network parameters are trained in the conventional way and the pruning mask is obtained using Eq. (4). Convolutional layers are pruned based on the obtained pruning mask, and fully connected layers are binarized based on Eq. (7). The quantization and pruning steps cause an accuracy drop which may be alleviated during retraining.

## IV. EXPERIMENTAL RESULTS

For the evaluation of the proposed method, experiments are performed using the TensorFlow framework. WCE bleeding images from publicly available databases are used [37-38]. Five bleeding images from [37], and 50 images from [38] are used for training and testing. To study the performance of the proposed method, we use two metrics. The first metric is DICE score defined as:

$$DICE = \frac{2TP}{2TP + FP + FN} \qquad (8)$$

where TP and TN are the numbers of pixels that are correctly classified as bleeding and non-bleeding, respectively. FP and FN are the numbers of pixels that are incorrectly classified as bleeding and non-bleeding, respectively. The second metric is *area under the curve* of ROC (AUC of ROC) which is used for classification performance. Patch size is selected to be 9×9 experimentally. Experiments are performed mainly on two neural network structures including MLP and CNN as followings.

*A. Imbalance data problem*

Experiments on our WCE bleeding image dataset show that only 0.2% of the pixels belong to the bleeding class and the rest are non-bleedings. This problem is known as imbalance data. Imbalance data problem can be solved in the training phase by selecting the proper ratio of bleeding and non-bleeding patches. As most of the pixels are non-bleeding, training the neural network with the original dataset would result in a tendency toward the non-bleeding class and the network would not be able to classify the bleeding regions correctly. In what follows, for each network structure, the ratio of bleeding and non-bleeding pixels is selected in a way that non-bleeding pixels do not overwhelm the bleeding pixels.

*B. Simplified MLP configuration*

Different MLP network configurations are tested, and the best architecture is selected. An MLP structure with three hidden layers consisted of 40, 20, and 8 neurons, respectively is applied as a simple structure for bleeding region segmentation. The sigmoid function is used as activation function, cross-entropy as loss function and Adam optimizer is used for optimization of the loss function. 500,000 patches are randomly selected in which the number of bleeding and the number of non-bleeding patches are equal.

*C. Simplified CNN configuration*

To evaluate the simplification method for CNN, a patch based CNN structure with 64 and 32 convolutional filters in each convolutional layer and fully connected layers of size 60 and 40 are used. The RELU activation function and pooling follow convolutional layers. Adam optimization method is used as an optimizer and cross-entropy as the loss function. The aforementioned configurations of the applied CNN are chosen by conducting numerous experiments using the same dataset as the MLP experiment to obtain the best results. Batch normalization and dropout are utilized for efficient training and dataset are balanced for better training. 600,000 patches are selected such that the bleeding patches are one-third of the normal patches.

Table II. Segmentation performance in term of DICE score

| | Method | DICE | AUC of ROC |
|---|---|---|---|
| [8] | SVM | 0.84 | -- |
| [9] | SVM | 0.81 | -- |
| [10] | SVM | 0.748 | -- |
| [14] | SVM | -- | 0.835 |
| [7] | SVM | 0.862 | -- |
| Our MLP (Quantized) | MLP-ANN | 0.831 | 0.974 |
| Our MLP (Full Precision) | MLP-ANN | 0.861 | 0.983 |
| Our CNN (Quantized) | CNN | 0.846 | 0.978 |
| CNN (Pruned - Quantized) | CNN | 0.869 | 0.985 |
| Our CNN (Full Precision) | CNN | 0.890 | 0.984 |

### D. Quantitative results

Different network structures and simplification methods are quantitatively compared in Table II. Also, results of other related works on the segmentation of the bleeding regions in WCE images are reported in Table II. Other works are based on different feature extraction methods and SVM for the classification. Since the number of normal pixels is much larger than the abnormal pixels, DICE score is used as the performance metric. For better comparison AUC of ROC metric is also included in Table II. It can be observed that our full precision CNN have DICE score equal to 0.89. Selection of proper image color channels, using a balanced dataset, and configuration of an efficient CNN are the main advantages of the proposed full-precision CNN structure. As illustrated in Table II, the simplified (quantized) MLP has about %3 drop in DICE score in comparison with the full precision MLP. Simplified MLP does not require any multiplications and has a simple structure in comparison with the CNN. If a more accurate model is demanded, it is possible to use a simplified CNN. From Table II, it can be observed that quantizing all of the CNN parameters leads to about %4.5 drop in DICE score which can be reduced to about %2.2 with pruned-quantized CNN. Pruned-quantized CNN and full precision MLP have nearly similar results; however their structural complexities are very different. For better insight into the complexity of different network structures, structural complexities of the employed networks are analyzed in subsection IV.F. Also, the AUC illustrated in Table II shows that both pruned-quantized and full-precision CNNs have better segmentation results than the other methods.

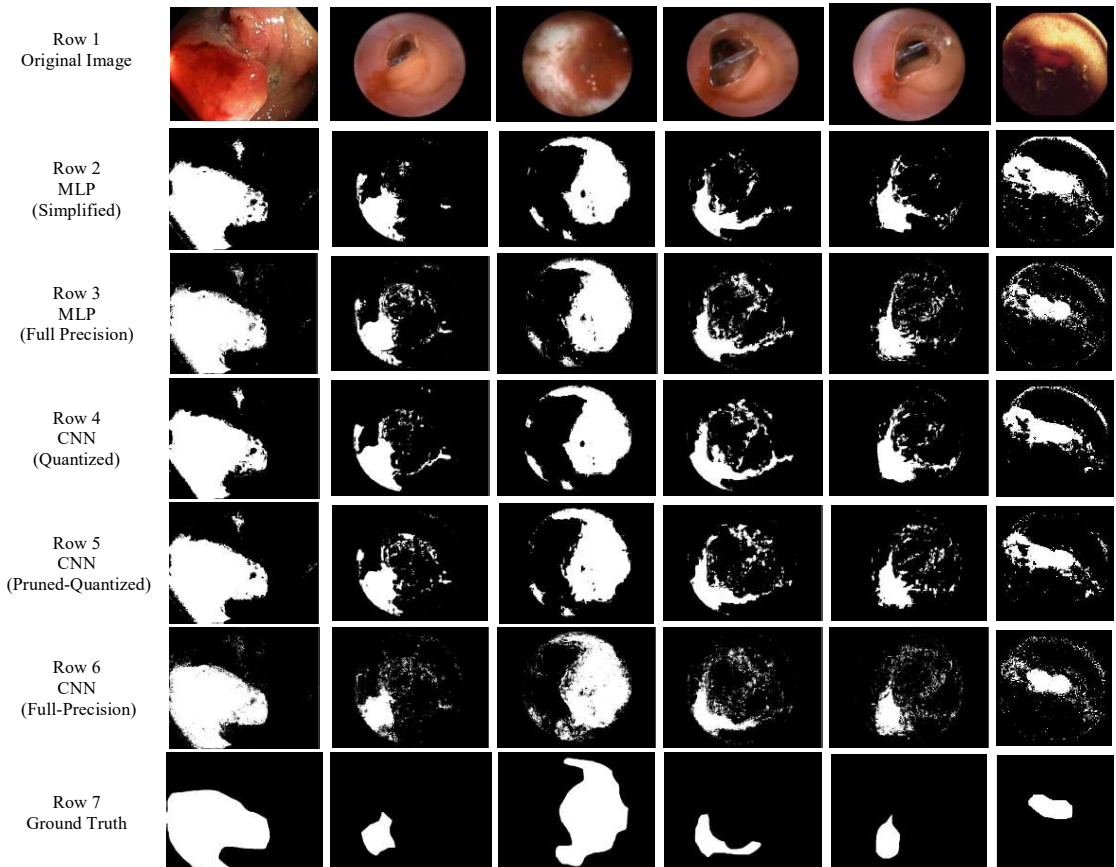

Figure 8. Visual results of different networks for bleeding segmentation

Table III. Summary of parameters in original and simplified MLP structure

| Layer | Type | Neurons | Original Weights | Simplified Weights |
|---|---|---|---|---|
| 1 | FC | 40 | 9720 | Quantized |
| 2 | FC | 20 | 800 | Quantized |
| 3 | FC | 8 | 160 | Quantized |
| 4 | FC | 2 | 16 | Quantized |

*E. Qualitative results*

In order to evaluate the proposed method visually, qualitative results of the full-precision and simplified MLP and the full-precision, quantized, and pruned-quantized CNN related to six sample images of the utilized dataset are provided in Fig. 8. Segmentation results are in the form of a binary mask, and the ground truth of this mask is shown in row 7 of Fig. 8. Visual results of the simplified and full precision MLP are illustrated in the second and third rows of Fig. 8, respectively. Both MLP networks have acceptable visual results, and simplified MLP has no significant degradation of the visual quality in comparison with its full precision version. Also, the visual results of quantized, pruned-quantized, and full precision of CNN are illustrated in rows 4, 5 and 6 of Fig. 8, respectively. CNN with full precision parameters have the best visual results comparing with results of other networks. It can be observed in row 4 of Fig. 9, that CNN with only quantization has created some visible artifact due to the quantization error. As an example, there are some small regions in the ground truth resulted from the fully quantized CNN that are wrongly predicted as bleeding regions but the original CNN predicted them correctly. This error is alleviated in the visual result of the pruned-quantized network due to the efficient simplification method. Generally, two simplified network structures are considered including simplified MLP and pruned-quantized CNN with visual results illustrated in row 2 and 5 of Figure 8. Pruned-quantized CNN network has better visual results, but the complexity of the simplified MLP network makes it suitable for implementation inside the capsule endoscopy device.

For further evaluation of the proposed method, ROC curves and AUC of ROC based on classification results of both MLP networks and all three versions of the CNN are presented in Fig. 9 and Fig. 10, respectively. MLP with original parameters (full-precision) and simplified MLP network have similar ROC curve in Fig. 9. Also in Fig. 10, the ROC curve of all CNN networks is similar to an AUC of about 0.98. Figure 10 shows that a CNN has an appropriate segmentation capability. It is worth mentioning that MLP with a very simple structure has an acceptable AUC of ROC.

*F. Complexity analysis*

Suppose that the designed networks are going to be implemented on a device with limited hardware resource such as the capsule device. In devices such as WCE, implementing an algorithm which requires a small number of computational operations is very beneficial. For both of the proposed simplified networks, the computational complexity is considered as the number of multiplications and their bit-length and is reported in Table III and Table IV, respectively. The number of parameters contained in each layer of both original and simplified MLP neural network is presented in Table III and also the number of parameters contained in each FCL and CL of all three versions of the utilized CNN is presented in Table IV.

The number of network biases is negligible and has no significant effect on the total computational complexity; hence biases are ignored in Table III and Table IV. In Table

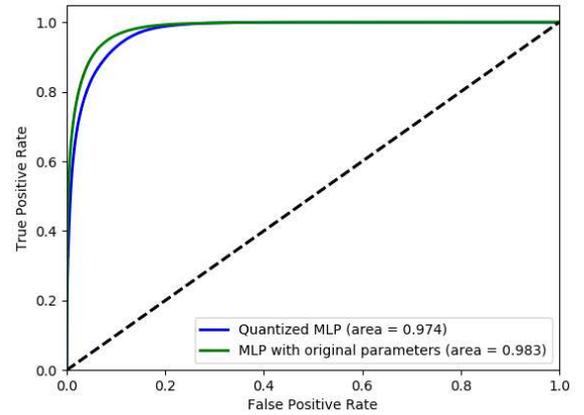

Figure 9. ROC of MLP segmentation networks

Table IV. Summary of parameters in original and simplified CNN structure

| Layer | Type | Maps and Neurons | Filter Size | Original Weights | Simplified Weights |
|---|---|---|---|---|---|
| 1 | Input | 1M × 9×9 | - | - | - |
| 2 | Convolution | 64M × 9×9 | 3×3 | 1728 | 979 |
| 3 | Max Pooling | 64M × 5×5 | 2×2 |  | - |
| 4 | Convolution | 32M × 5×5 | 3×3 | 18432 | 9163 |
| 5 | Max Pooling | 32M × 3×3 | 2×2 | - | - |
| 6 | FC | 60 | 1×1 | 17280 | Quantized |
| 7 | FC | 40 | 1×1 | 2400 | Quantized |
| 8 | FC | 2 | 1×1 | 80 | Quantized |

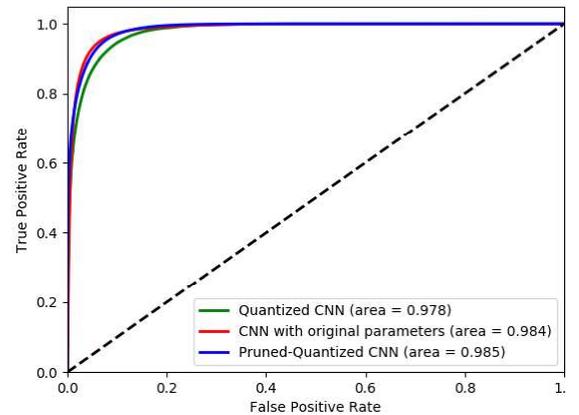

Figure 10. ROC of CNN segmentation networks

III, for the MLP structure, all 32-bit multiplication operations are converted to 1-bit operations using quantization. 1-bit operations can be realized by addition without any multiplications that simplifies the computations needed by the network.

Also, the types of layers in the utilized CNN are illustrated in Table IV. The number of parameters in full precision and simplified (pruned-quantized) CNN are compared. It is observed that the number of 32-bit length weights in full-precision CNN is reduced from 1728 to 979 in the first CL and from 18432 to 9163 in the second CL. Also, all of the weights in the FCLs are binarized.

Finally, both of the designed networks are compared with each other concerning their complexity in Table V. In Table V the total number of employed parameters in term of 32-bit, and 1-bit length operations are reported. Simplified CNN (pruned-quantized CNN) have 10142 thirty-two-bit and 19760 1-bit weights. Although CNN has better DICE score, the simplified MLP uses only 10696 1-bit weights. Binary operations reduce the complexity of hardware implementation of segmentation significantly and make segmentation process suitable for implementing inside the capsule device.

Table V. Number of parameters in two simplified networks

| Method | # of parameters | |
| --- | --- | --- |
| | 32-bit | 1-bit |
| Our simplified CNN | ~10,142 | ~19760 |
| Our simplified MLP | 0 | ~10696 |

## V. CONCLUSION

In this paper, simplification methods for reducing the structural complexity of the neural networks for automatic detection of bleeding regions in WCE images were investigated. Two neural networks including MLP and CNN were applied for segmentation as different variations of neural networks' structure. The detection system was designed concerning limitation of the hardware resources in the WCE device. Hence, the MLP with quantized weights was utilized, and a new method based on pruning and quantization was employed for the CNN. In comparison with the original CNN, the pruned-quantized CNN needed almost 50% fewer parameters in CL, and also each of the 19760 weights in FCL was quantized. Furthermore, simulation results showed that the quantized MLP with 10696 weights has comparable results with the simplified CNN. However, it needed much fewer parameters than the simplified CNN. Quantized neural networks without any multiplication could be considered as an automatic diagnostic approach inside the WCE device.

REFERENCES

[1] M. Keuchel and N. Kurniawan, *Video Capsule Endoscopy: A Reference Guide and Atlas*, vol. 20, no. 5. 2014.
[2] D. K. Iakovidis and A. Koulaouzidis, "Software for enhanced video capsule endoscopy: Challenges for essential progress," *Nat. Rev. Gastroenterol. Hepatol.*, vol. 12, no. 3, pp. 172–186, 2015.
[3] Y. Yuan, J. Wang, B. Li, and M. Q. H. Meng, "Saliency based ulcer detection for wireless capsule endoscopy diagnosis," *IEEE Trans. Med. Imaging*, vol. 34, no. 10, pp. 2046–2057, 2015.
[4] I. Mehmood, M. Sajjad, and S. W. Baik, "Video summarization based tele-endoscopy: A service to efficiently manage visual data generated during wireless capsule endoscopy procedure," *J. Med. Syst.*, vol. 38, no. 9, 2014.
[5] Y. Fu, W. Zhang, M. Mandal, and M. Q. H. Meng, "Computer-aided bleeding detection in WCE video," *IEEE J. Biomed. Heal. Informatics*, vol. 18, no. 2, pp. 636–642, 2014.
[6] S. Suman, F. A. B. Hussin, A. S. Malik, K. Pogorelc, M. Riegler, S. H. Ho, I. Hilmi, and K. L. Goh, "Detection and classification of bleeding region in wce images using color feature," in *Proceedings of the 15th International Workshop on Content-Based Multimedia Indexing - CBMI '17*, 2017, pp. 1–6.
[7] T. Ghosh, S. A. Fattah, K. A. Wahid, W.-P. Zhu, and M. O. Ahmad, "Cluster based statistical feature extraction method for automatic bleeding detection in wireless capsule endoscopy video," *Comput. Biol. Med.*, 2018.
[8] E. Tuba, M. Tuba, and R. Jovanovic, "An algorithm for automated segmentation for bleeding detection in endoscopic images," *Proc. Int. Jt. Conf. Neural Networks*, vol. 2017–May, pp. 4579–4586, 2017.
[9] F. Deeba, F. M. Bui, and K. A. Wahid, "Automated Growcut for segmentation of endoscopic images," *Proc. Int. Jt. Conf. Neural Networks*, vol. 2016–Oct., pp. 4650–4657, 2016.
[10] Y. Yuan, B. Li, and M. Q. H. Meng, "Bleeding frame and region detection in the wireless capsule endoscopy video," *IEEE J. Biomed. Heal. Informatics*, vol. 20, no. 2, pp. 624–630, 2016.
[11] A. Karargyris and N. Bourbakis, "Detection of small bowel polyps and ulcers in wireless capsule endoscopy videos," *IEEE Trans. Biomed. Eng.*, vol. 58, no. 10 PART 1, pp. 2777–2786, 2011.
[12] Y. Yuan, B. Li, and M. Q. H. Meng, "Improved bag of feature for automatic polyp detection in wireless capsule endoscopy images," *IEEE Trans. Autom. Sci. Eng.*, vol. 13, no. 2, pp. 529–535, 2016.
[13] Y. Yuan, X. Yao, J. Han, L. Guo, and M. Q. H. Meng, "Discriminative joint-feature topic model with dual constraints for WCE classification," *IEEE Trans. Cybern.*, pp. 1–12, 2017.
[14] D. K. Iakovidis and A. Koulaouzidis, "Automatic lesion detection in wireless capsule endoscopy - A simple solution for a complex problem," *Image Process. (ICIP), 2014 IEEE Int. Conf. on. IEEE*, pp. 2236–2240, 2014.
[15] X. Jia and M. Q. H. Meng, "Gastrointestinal bleeding detection in wireless capsule endoscopy images using handcrafted and CNN features," *Proc. Annu. Int. Conf. IEEE Eng. Med. Biol. Soc. EMBS*, pp. 3154–3157, 2017.
[16] X. Jia and M. Q.-H. Meng, "A deep convolutional neural network for bleeding detection in wireless capsule endoscopy images," *2016 38th Annu. Int. Conf. IEEE Eng. Med. Biol. Soc.*, pp. 639–642, 2016.
[17] X. Jia and M. Q.-H. Meng, "A study on automated segmentation of blood regions in wireless capsule endoscopy images using fully convolutional networks," in *Biomedical Imaging (ISBI 2017), 2017 IEEE 14th International Symposium on*, 2017, pp. 179–182.
[18] A. K. Sekuboyina, S. T. Devarakonda, and C. S. Seelamantula, "A convolutional neural network approach for abnormality detection in Wireless Capsule Endoscopy," *2017 IEEE 14th Int. Symp. Biomed. Imaging (ISBI 2017)*, pp. 1057–1060, 2017.
[19] J. Long, E. Shelhamer, and T. Darrell, "Fully convolutional networks for semantic segmentation," in *Proceedings of the IEEE Computer Society Conference on Computer Vision and Pattern Recognition*, 2015, vol. 07–12–June, pp. 3431–3440.
[20] M. A. Khorsandi, N. Karimi, S. Samavi, M. Hajabdollahi, S. M. R. Soroushmehr, K. Ward, and K. Najarian, "Hardware image assessment for wireless endoscopy capsules," pp. 2050–2053, 2016.
[21] P. Turcza and M. Duplaga, "Hardware-efficient low-power image processing system for wireless capsule endoscopy," *Biomed. Heal. Informatics, IEEE J.*, vol. 17, no. 6, pp. 1046–1056, 2013.
[22] P. Turcza and M. Duplaga, "Sensors and actuators a : physical low power FPGA-based image processing core for wireless capsule endoscopy," *Sensors Actuators A. Phys.*, vol. 172, no. 2, pp. 552–560, 2011.
[23] N. Jarray, M. Elhaji, and A. Zitouni, "Low complexity and efficient architecture of 1D-DCT based Cordic-Loeffler for wireless endoscopy capsule," in *12th International Multi-Conference on Systems, Signals and Devices, SSD 2015*, 2015.

[24] J. H. Tan, H. Fujita, S. Sivaprasad, S. V. Bhandary, A. K. Rao, K. C. Chua, and U. R. Acharya, "Automated segmentation of exudates, hemorrhages, microaneurysms using single convolutional neural network," *Inf. Sci. (Ny).*, vol. 420, pp. 66–76, 2017.

[25] M. Hajabdollahi, N. Karimi, S. M. Reza Soroushmehr, S. Samavi, and K. Najarian, "Retinal blood vessel segmentation for macula detachment surgery monitoring instruments," *Int. J. circuit theory Appl.*, 2018.

[26] E. Nasr-Esfahani, N. Karimi, M. H. Jafari, S. M. R. Soroushmehr, S. Samavi, B. K. Nallamothu, and K. Najarian, "Segmentation of vessels in angiograms using convolutional neural networks," *Biomed. Signal Process. Control*, 2018.

[27] M. Courbariaux, I. Hubara, D. Soudry, R. El-Yaniv, and Y. Bengio, "Binarized neural networks: Training neural networks with weights and activations constrained to +1 or −1," *arXiv*, p. 9, 2016.

[28] M. Courbariaux, Y. Bengio, and J.-P. David, "Binary connect: Training deep neural networks with binary weights during propagations," in *Advances in neural information processing systems*, 2015, pp. 3123–3131.

[29] M. Horowitz, "1.1 Computing's energy problem (and what we can do about it)," *Dig. Tech. Pap. - IEEE Int. Solid-State Circuits Conf.*, vol. 57, pp. 10–14, 2014.

[30] S. J. Hanson and L. Pratt, "Comparing biases for minimal network construction with back-propagation," *Adv. neural Inf. Process. Syst. 1*, pp. 177–185, 1989.

[31] Y. Le Cun, J. S. Denker, and S. a Solla, "Optimal brain damage," *Adv. Neural Inf. Process. Syst.*, vol. 2, no. 1, pp. 598–605, 1990.

[32] S. Han, J. Pool, J. Tran, and W. Dally, "Learning both weights and connections for efficient neural network," in *Advances in neural information processing systems*, 2015, pp. 1135–1143.

[33] H. Li, A. Kadav, I. Durdanovic, H. Samet, and H. P. Graf, "Pruning filters for efficient convnets," *arXiv Prepr. arXiv1608.08710*, 2016.

[34] M. S. Miri and A. Mahloojifar, "Retinal image analysis using curvelet transform and multistructure elements morphology by reconstruction," *IEEE Trans. Biomed. Eng.*, 2011.

[35] R. Lukac and K. N. Plataniotis, *Color image processing: methods and applications*. CRC press, 2006.

[36] R. M. Gray, *Entropy and information theory*. Springer Science & Business Media, 2011.

[37] A. Koulaouzidis, "KID: Koulaouzidis-Iakovidis database for capsule endoscopy." 2016.

[38] F. Deeba, "Bleeding images and corresponding ground truth of CE images," 2016. Available online at https: //sites.google .com/site /farahdeeba073/Research/resources, accessed on March 2017.